\documentclass[10pt, conference, compsocconf]{IEEEtran}
\usepackage{amsmath,graphicx}
\usepackage{times}
\usepackage{epsfig}
\usepackage{graphicx}
\usepackage{amsmath}
\usepackage{amssymb}
\usepackage{array}
\usepackage{multirow}
\usepackage{booktabs}
\usepackage{cite}

\ifCLASSINFOpdf
\else
\fi
\begin{document}
%
\title{OSVNet: Convolutional Siamese Network for Writer Independent Online Signature Verification}





%
\author{\IEEEauthorblockN{Chandra Sekhar \IEEEauthorrefmark{1},
Prerana Mukherjee \IEEEauthorrefmark{1},
Devanur S Guru\IEEEauthorrefmark{2} and
Viswanath Pulabaigari\IEEEauthorrefmark{1} }
\IEEEauthorblockA{\IEEEauthorrefmark{1}Indian Institute of Information Technology, Sri City, Andhra Pradesh
\\ Email: \{chandrasekhar.v, prerana.m, viswanath.p\}@iiits.in}
\IEEEauthorblockA{\IEEEauthorrefmark{2}University of Mysore\\
Email: dsg@compsci.uni-mysore.ac.in}}


\maketitle

\begin{abstract}
Online signature verification (OSV) is one of the most challenging tasks in writer identification and digital forensics. Owing to the large intra-individual variability, there is a critical requirement to accurately learn the intra-personal variations of the signature to achieve higher classification accuracy. To achieve this, in this paper, we propose an OSV framework based on deep convolutional Siamese network (DCSN). DCSN automatically extracts robust feature descriptions based on metric-based loss function which decreases intra-writer variability (Genuine-Genuine) and increases inter-individual variability (Genuine-Forgery) and directs the DCSN for effective discriminative representation learning for online signatures and extend it for one shot learning framework. Comprehensive experimentation conducted on three widely accepted benchmark datasets MCYT-100 (DB1), MCYT-330 (DB2)
and SVC-2004-Task2 demonstrate the capability of our framework to distinguish the genuine and forgery samples. Experimental results confirm the efficiency of deep convolutional Siamese network based OSV by achieving a lower error rate as compared to many recent and state-of-the art OSV techniques.
\end{abstract}


%
\IEEEpeerreviewmaketitle

\section{Introduction}
\label{sec:introduction}
Biometrics is an automated approach of person identification and verification that are based on personal physiological features like human gait, iris, fingerprints and the structure of the retina, veins etc. or based on personal behavioral traits such as signature, hand writing, key stroke dynamics etc. \cite{guru2017interval}. Among these biometric modalities, due to cost-effective acquisition and resistance to physical tamper, online signature is the most popular technique for person identification in polymorphous m-commerce and m-payment applications \cite{lai2017online}. Online signature is defined by real time signals varying over time, in which the dynamic features are acquired through specialized devices like Graphic Tablets, Stylus Pens etc. which enables reading both the structural information ($x, y$ coordinates) and the dynamic properties such as inclination, velocity, pressure, acceleration of a pen as it marks out its successive points \cite{guru2017interval, pascanu2013difficulty,li2017classification}.  

In literature many online signature verification (OSV) frameworks have been proposed which can be broadly classified into feature-based methods \cite{guru2017interval, pirlo2015multidomain, manjunatha2016online, sharma2018exploration} that analyze signatures based on a set of global or local features, function-based methods which employ various techniques like feature fusion based \cite{guru2009online}, Hidden Markov models \cite{maiorana2010cancelable}, DTW 
\cite{liu2015online, sharma2018exploration, parziale2019sm, sharma2016enhanced, lai2017online}, matching based \cite{tang2018information},  divergence based \cite{tang2018information}, neural network based \cite{tolosana2017biometric}, Gaussian Mixture Models \cite{alaei2017efficient, sharma2017novel}, random forest \cite{sharma2017novel}, sequence matching \cite{tang2018information} , stability based \cite{parziale2019sm}, Deep learning based \cite{lai2017online, lai2018recurrent} etc. 

Recently, the work by \cite{tolosana2017biometric, lai2017online, lai2018recurrent} on Recurrent Neural Networks (RNN’s) has proven to be very efficient in recognising and modelling hidden patterns in time series data 
by learning relationship that exists between current inputs and past data. Hence, RNN based frameworks are widely used in financial markets, speech signals, OSV etc.\cite{tolosana2017biometric,lai2018recurrent }. However, the traditional RNNs suffer from an inherent drawback of vanishing gradients or exploding gradients during the backpropagation step of training process with the long input sequence \cite{pascanu2013difficulty}. In addition to these drawbacks, the framework based on LSTM RNN architecture should be trained with both the genuine and forgery samples every time a new user is enrolled into the system. Getting the forgery samples upfront may not be feasible in real time scenarios \cite{tolosana2017biometric}. 

In such scenarios, an online signature verification can be efficiently modelled by Siamese networks \cite{lai2018recurrent} which consists of twin convolutional networks accepting two distinct online signatures and learning a similarity metric from pairs of signatures (through powerful discriminative features) which decreases intra-writer variability i.e. pairs of signatures from the same user (genuine-genuine) and increase the inter-individual variability i.e. pairs of signatures from different people (genuine-forgery). As the network is learning a similarity metric rather learning the features from the training samples, the model can be generalized to classify the signatures from unknown users without providing forgery training signature samples \cite{li2017classification}. In addition to the abovementioned motivation, sharing weights across subnetworks results in less parameters (weights and biases) to train, which in turn resists the model tendency to over-fit. 

Even though Siamese networks overcome the drawbacks of traditional RNN and LSTM based frameworks, and have great scope of applicability in online signature verification, very few studies \cite{tolosana2017biometric, lai2017online, lai2018recurrent, ahrabian2017usage} have been reported on application of LSTM RNNs to online signature verification. 

Therefore, this paper focuses on the most challenging co-variate of online signature verification. (i) To the best of author's knowledge, this is the first work in which we propose a Siamese based online signature verification (OSV) framework using CNNs, which enables one-shot learning for online signature verification tasks, resulting in substantial reduction of the parameter count and the amount of computation required. ii) Extensive experiments validate that the proposed framework has better performance on the benchmark datasets over the state of the art online signature verification techniques.

The manuscript is organized as follows. In Sec. \ref{sec:proposed_method}, we discuss about our proposed OSVnet architecture. In Sec. \ref{sec:results}, we provide details of the training and testing data, experimental analysis along with the results and comparison of the proposed framework with the recent state of the art baseline models are discussed. In Sec. \ref{sec:conclusion}, we provide the conclusion and future work.

\section{PROPOSED FRAMEWORK FOR SIGNATURE VERIFICATION}
\label{sec:proposed_method}
In this section, we describe the proposed Siamese network based OSV framework in detail. 
\subsection{Input Signature Format}
\begin{figure}[h]
       \centering
       \fbox{
       \includegraphics[scale=0.45]{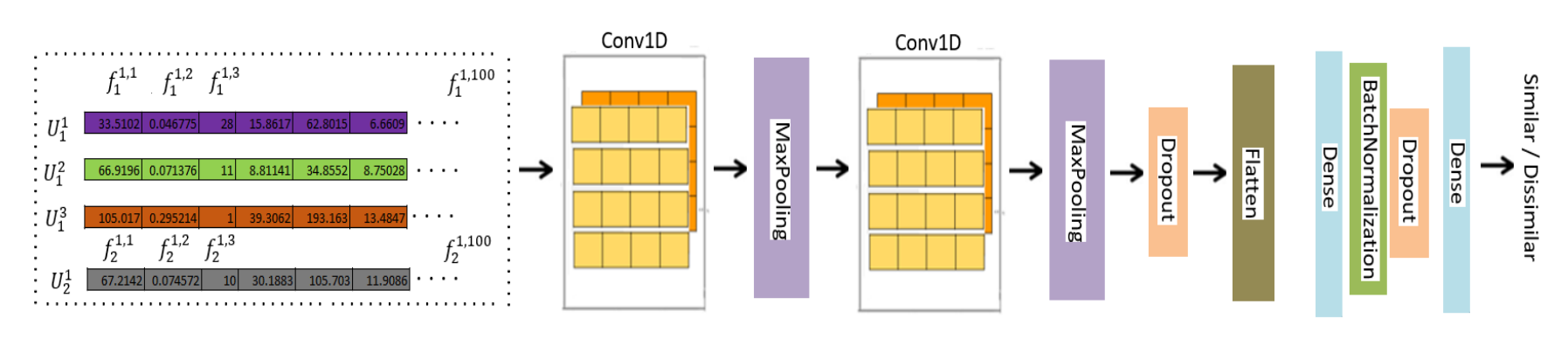}}
       \caption{Overview of the  CNN architecture of the Proposed OSV framework.}
       \label{fig:overview}
\end{figure}

\begin{figure}[h]
       \centering
       \fbox{
       \includegraphics[scale=0.25]{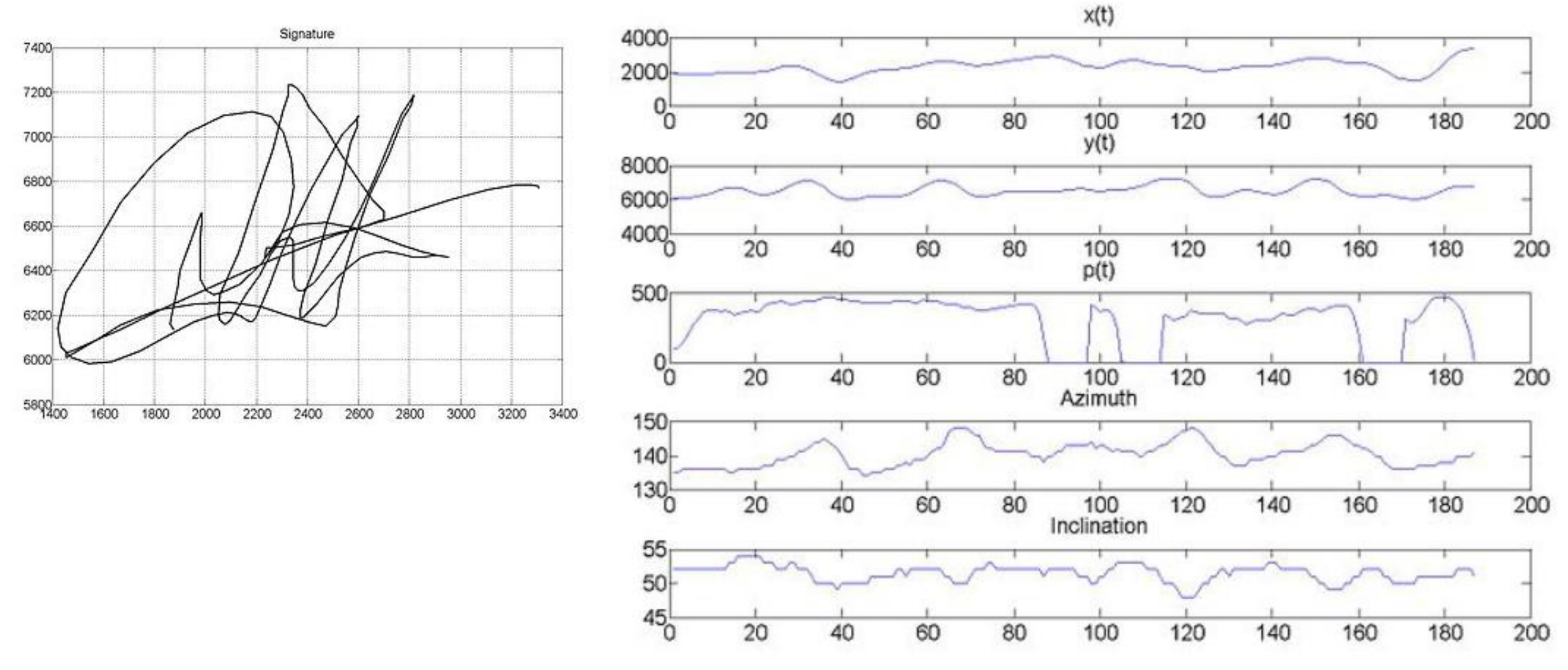}}
       \caption{A sample online signature from the
MCYT-100 signature corpus.}
       \label{fig:signature}
\end{figure}

As depicted in Fig. \ref{fig:overview} and \ref{fig:sample}, the input to the framework is an online signature. An online signature is a row vector of dimensionality $1$ x $100$ in case of MCYT-100 dataset and $1$ x $47$ in case of SVC dataset. Values $100$, $47$ represent the total number of global features computed for each writer’s signature. The local features like ($x$-coordinate, $y$-coordinate, pressure, Azimuthal angle) are extracted at each of point of signature as shown in Fig. \ref{fig:signature} and these extracted local features are used to compute the global features to represent the user signature e.g. max velocity, average pressure, standard deviation of acceleration etc. \cite{guru2007symbolic, guru2009online}. 

\subsection{CNN and Siamese Network}
\begin{figure*}[h]
       \centering
       \fbox{
       \includegraphics[scale=0.75]{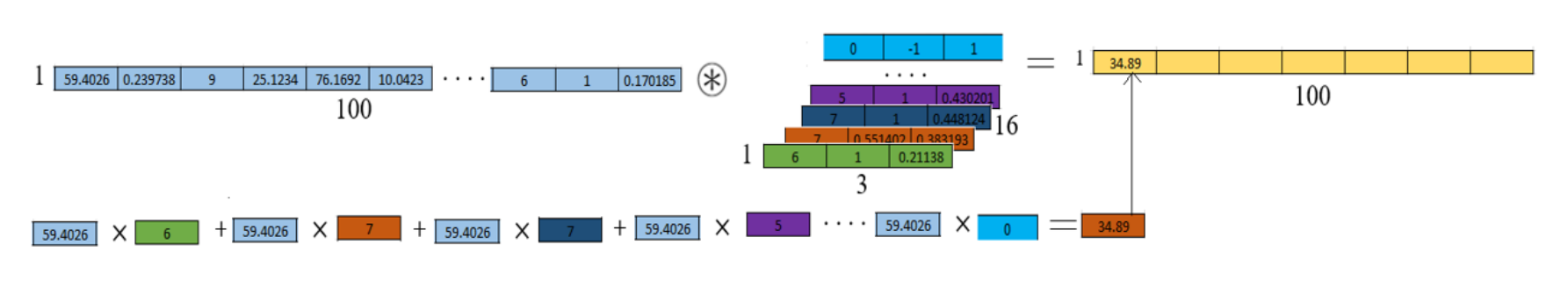}}
       \caption{A sample demonstration of convolution operation between online signature and the kernels.}
       \label{fig:sample}
\end{figure*}

The architecture of the CNN layers is depicted in Fig. \ref{fig:overview}. Deep Convolutional Neural Networks (CNN) are collection of several convolutional and pooling layers. Kernel of different size perform convolution operation on the input signature and outputs the feature maps. The feature maps form an input to the pooling layers, which down samples the feature maps before feeding to higher level layers. As online signature is a one-dimensional vector as shown in Fig. \ref{fig:sample}, one dimensional convolution operation is performed between the input signature and the one-dimensional kernel. We have used $16$ kernels of size $1$x$3$ to convolve on the input signature. 
\begin{figure*}[h]
       \centering
       \fbox{
       \includegraphics[scale=0.65]{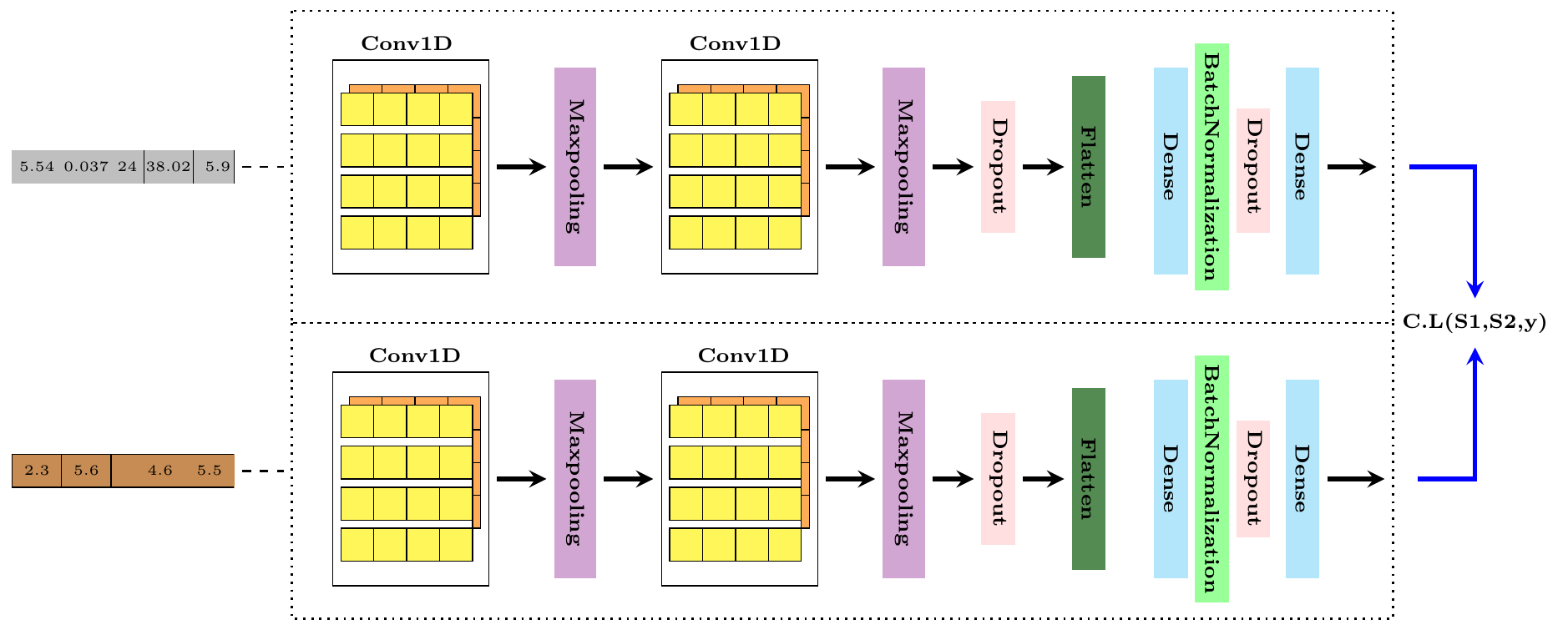}}
       \caption{Architecture of proposed Siamese based OSVNet framework.}
       \label{fig:archi}
\end{figure*}

As depicted in Fig. \ref{fig:archi}, the Siamese network is a collection of twin convolutional neural network with the shared weights and biases. Siamese networks have been successfully used in real time applications like Real-Time Object Tracking \cite{he2018twofold}, Real time visual tracking \cite{li2017classification} etc. The parameters updated in one CNN networks will reflect in second network also. As depicted in Fig. \ref{fig:archi}, a pair of signatures forms an input to the twin CNNs and a series of convolution and pooling operations are performed on the input signatures and finally a high-level feature representation are learnt from each network. These feature representations are joined by a most widely used contrastive loss function \cite{tolosana2017biometric, lai2017online, lai2018recurrent}, which inherently computes the Euclidean distance between them and learns the similarity metric. The contrastive loss which is a margin-based loss function can be described as follows,
\begin{equation}
  CL(S1,S2,y)=y* \left \|S1,S2  \right \|^2+(1-y)*max⁡\left (0,m^2 -\left \| S1,S2 \right \| ^2\right ), 
  \label{eq:eq1}
\end{equation}

where $S1$, $S2$ are signature samples, `$y$' is a binary value, which indicates whether the input samples are in proximity or not. `$m$' is the margin value, in our case and is equal to 1. $\left \|S1,S2  \right \|^2$ represents the Euclidean distance between two samples. Euclidean distance is computed in the embedded feature space using an embedding function `$f$' that maps a signature feature vector to real vector space through CNN. Unlike traditional CNNs networks which learns an approximate function to classify the input signature samples into binary cases i.e. genuine or forgery, Siamese network aims to learn the similarity metric which minimizes the output feature representations for input signature pairs that are genuine, and maximizes the feature representations if the input signature samples are genuine-forgery category.
\subsection{CNN and Siamese Architecture}

We have used a CNN architecture that is inspired by Yilmarz et al. \cite{yilmaz2018hybrid} which was developed for an offline signature verification problem. We have modified the architecture to suit for online signature, which is of one dimension. For the reproducibility of our results, in Table \ref{tab:tab1}, we have listed all the parameters used in designing the CNN network. For convolution and pooling layers, we use the notation $N$ x $H$ x $W$ to represent the number of kernels, height and width of the particular kernel. In the framework, stride signifies the distance between the current and next location of kernel to perform the convolution operation. To make the CNN to approximate the complex functions and to induce non-linearity, we have used `\textit{ReLu}' as an activation function. In case of fully connected layers we have used `\textit{Sigmoid}' as an activation function. To normalize the feature representations from both the CNNs, we have used Local Response Normalization technique as discussed in \cite{krizhevsky2012imagenet}. To resist the model to become overfit and to make the framework to learn the hyperparameters rather than memorizing the output, we have used Dropout of 50\% each, one after the second max pooling layer, and the second one after the batch normalization layer.
\begin{table*}[]
\scriptsize
\caption{OVERVIEW OF THE CONSTITUTING CNNS AND TRAINING HYPER-PARAMETERS. `-' represents that the value is not applicable.}
\begin{center}
\begin{tabular}{ccc||cc}
\hline
\textbf{Layer}      & \textbf{Size} & \textbf{Parameters} & \textbf{Attribute}                 & \textbf{Value}                                                      \\ \hline \hline
Convolution         & 16x1x100      & padding=`same'      & Initializer Function               & -                                                                   \\ 
Pooling             & 50x16         & -                   & Activation Function                & Relu                                                                \\
Convolution         & 16x1x50       & padding=`same'      & Mini Batch Size                    & 36                                                                  \\ 
Pooling             & 25x16         & padding=`same'      & Loss Function                      & Binary crossentropy                                                 \\ 
Dropout             & -             & 0.5                 & Optimizer                          & Adam                                                                \\ 
Dense               & 36            & -                   & $beta_1$, $beta_2$, epsilon, decay & \begin{tabular}[c]{@{}c@{}}0.9, 0.999, 1e-08,\\   0.00\end{tabular} \\
Batch Normalization & 36            & -                   & Early Stopping                     & Patience = 5, Min$\Delta$ = 0                                       \\ 
Dropout             & -             & 0.5                 & Learning rate                      & 0.004                                                               \\ 
Dense               & 36            & -                   & Epochs                             & 400                                                                 \\ 
-                   & -             & -                   & Bias initializer                   & $random_{uniform}$                                                    \\ 
-                   & -             & -                   & Depthwise initializer              & $random_{uniform}$                                                    \\ 
-                   & -             & -                   & Kernel initializer                 & $random_{uniform}$                                                    \\ 
-                   & -             & -                   & Kernel constraint                  & $max_{norm}(4)$                                                       \\ 
-                   & -             & -                   & Bias constraint                    & $max_{norm}(4)$                                                       \\
-                   & -             & -                   & Kernel regularizer                 & $regularizers.l2(0.03)$                                             \\ \
-                   & -             & -                   & Bias regularizer                   & $l2(0.03)$                                                          \\ \hline
\end{tabular}
\end{center}
\label{tab:tab1}
\end{table*}

\begin{table*}[]
\scriptsize
\caption{DATASET DETAILS USED IN THE EXPERIMENTS FOR THE PROPOSED FRAMEWORK.}
\begin{center}
\begin{tabular}{cccc}
\hline
\textbf{Dataset$\rightarrow$}                                                                          & \textbf{MCYT-100} & \textbf{MCYT-300}  & \textbf{SVC}   \\ \hline \hline
\textbf{Total number of Users}                                                                         & 100               & 330                & 40             \\ \hline
\textbf{Total number of features}                                                                      & 100               & 100                & 47             \\ \hline
\textbf{Number of genuine signatures per user}                                                         & 25                & 25                 & 20             \\ \hline
\textbf{Number of (genuine+genuine) combinations per user}                                             & $25_{C_2}$=300      & $25_{C_2}$=300       & $20_{C_2}$=200   \\ \hline
\textbf{\begin{tabular}[c]{@{}c@{}}Total number of \\ (genuine+genuine) combinations\end{tabular}}     & 300*100= 30000    & 300*330 = 99000    & 40*20 = 8000   \\ \hline
\textbf{Number of forgery signatures per user}                                                         & 25                & 25                 & 20             \\ \hline
\textbf{\begin{tabular}[c]{@{}c@{}}Number of\\   (genuine+forgery) combinations per user\end{tabular}} & 25 * 24 = 600     & 25 * 24 = 600      & 20 * 19 = 380  \\ \hline
\textbf{\begin{tabular}[c]{@{}c@{}}Total number of \\ (genuine+forgery) combinations\end{tabular}}     & 600 * 100 = 60000 & 600 * 330 = 198000 & 380*40 = 15200 \\ \hline
\textbf{Total number of genuine signatures}                                                            & 2500              & 8250               & 800            \\ \hline
\textbf{Total number of forgery signatures}                                                            & 2500              & 8250               & 800            \\ \hline
\textbf{Total Number of Samples}                                                                       & 5000              & 16500              & 1600           \\ \hline
\end{tabular}
\end{center}
\label{tab:tab2}
\end{table*}

As depicted in Fig. \ref{fig:overview}, our proposed framework is composed of four layers. The first two layers constitute the convolutional part of the CNN and are made up of two consecutive combinations of convolutional and max pooling layers. The input to the first convolution layer is an online signature of size $1$×$100$. The convolution layer use $16$ kernel of size $1$×$3$ to produce feature map of size $1$×$100$. We have applied one dimensional max pooling operation with $pool_{size}$ = 2 on the output of the first convolution layer, which results in down sampling of the feature map to $1$x$50$. The output from the second convolution layer forms an input to the one-dimensional max pooling layer, which results in the feature map of size $1$x$25$. Flatten reshapes the feature map of size $1$x$25$x$16$ into a one-dimensional feature vector of size $1$x$400$. The final dense layer results into a high level feature vector of size $1$x$36$ from each CNN of the Siamese network. These high level feature representations forms an input to the contrastive loss function described in Eq. \ref{eq:eq1}.

\section{EXPERIMENTATION AND RESULTS}
\label{sec:results}
The training parameters are presented in Table \ref{tab:tab1}. We have implemented our framework in Keras library with TensorFlow as backend. We have conducted our experiments on Nvidia, Titan X Pascal 12 GB GPU. We have extensively conducted verification experiments and validated the proposed Siamese based OSV framework by conducting the experiments on two widely accepted datasets i.e. MCYT-100 signature sub-corpus dataset (DB1) \cite{maiorana2010cancelable, diaz2018dynamic}, MCYT-330 signature sub corpus dataset (DB2) and SVC - Task 2 \cite{sharma2018exploration, kar2018stroke, yang2018online}. We detail the results in the following subsections.

\subsection{Experimental Protocol}
In this section we briefly discuss the experimentation and evaluation of the proposed Siamese based online signature framework. In order to evaluate the efficiency of our framework, we have conducted experiments on three widely used publicly available online signature benchmark datasets, viz., (1) MCYT-100, (2) MCYT-330, and (3) SVC -2004-Task2. A complete description of each dataset with respect to the experimental analysis is given in Table \ref{tab:tab2}.
The proposed Siamese based OSV framework is writer independent. To validate the writer independence, we split each dataset as follows. As depicted in Table \ref{tab:tab3}-\ref{tab:tab7}, we randomly select '$K$' users from a total of `$M$' users. `$K$' starts from 1 and gradually reaches ($M$-1). For each user $i$, where $1\leq i\leq  K$, we use the genuine and genuine combination as similar pairs, genuine and forgery combination as dissimilar pairs for training and the genuine and genuine, genuine and forgery combination of remaining  ($M-K$) users to test the accuracy of the framework. As depicted in Table \ref{tab:tab2}, for each user there are 300, 300, 200 possible genuine and genuine combinations and 600,600,380 genuine and forgery combinations are available in case of MCYT-100, MCYT-330 and SVC respectively. MCYT-330 dataset results in largest possible genuine and genuine combinations i.e. 99000 and 198000 genuine and forgery combinations. 
To overcome the class imbalance problem, we have selected equal number of genuine and genuine, genuine and forgery combinations. This results in effective training of the framework and eliminates the problem of over-fitting. 
\subsection{Results and Discussions}

\begin{table*}[]
\scriptsize
\caption{CLASSIFICATION ACCURACY OF THE PROPOSED FRAMEWORK WITH MCYT-100 DATASET INCLUDING FORGERY SAMPLES.}
\begin{center}
\begin{tabular}{ccccc}
\hline
\textbf{\begin{tabular}[c]{@{}c@{}}Number of Users for training  \\  (seen data)\end{tabular}} & \textbf{\begin{tabular}[c]{@{}c@{}}Number of Users for testing  \\ (unseen data)\end{tabular}} & \textbf{\begin{tabular}[c]{@{}c@{}}Number of Training\\ Signature Samples\end{tabular}} & \textbf{\begin{tabular}[c]{@{}c@{}}Number of Testing\\ Signature Samples\end{tabular}} & \textbf{Accuracy(\%)} \\ \hline \hline
95                                                                                           & 05                                                                                            & 57000                                                                                   & 3000                                                                                   & 93.90                 \\ 
90                                                                                           & 10                                                                                            & 54000                                                                                   & 6000                                                                                   & 93.02                 \\ 
80                                                                                           & 20                                                                                            & 48000                                                                                   & 12000                                                                                  & 92.85                 \\ 
70                                                                                           & 30                                                                                            & 42000                                                                                   & 18000                                                                                  & 92.78                 \\ 
60                                                                                           & 40                                                                                            & 36000                                                                                   & 24000                                                                                  & 91.79                 \\ 
50                                                                                           & 50                                                                                            & 30000                                                                                   & 30000                                                                                  & 90.48                 \\ 
40                                                                                           & 60                                                                                            & 24000                                                                                   & 36000                                                                                  & 91.46                 \\ 
30                                                                                           & 70                                                                                            & 18000                                                                                   & 42000                                                                                  & 92.48                 \\ 
20                                                                                           & 80                                                                                            & 12000                                                                                   & 48000                                                                                  & 92.85                 \\ 
10                                                                                           & 90                                                                                            & 6000                                                                                    & 54000                                                                                  & 86.55                 \\ 
05                                                                                           & 95                                                                                            & 3000                                                                                    & 57000                                                                                  & 85.02                 \\ 
01 (One Shot Learning)                                                                       & 99                                                                                            & 600                                                                                     & 59400                                                                                  & 78.16                 \\ \hline
\end{tabular}
\end{center}
\label{tab:tab3}
\end{table*}

\begin{table*}[]
\scriptsize
\caption{CLASSIFICATION ACCURACY OF THE PROPOSED FRAMEWORK  WITH MCYT-100 DATASET EXCLUDING FORGERY SAMPLES.}
\begin{center}
\begin{tabular}{ccccc}
\hline
\textbf{\begin{tabular}[c]{@{}c@{}}Number of Users for training \\ (seen data)\end{tabular}} & \textbf{\begin{tabular}[c]{@{}c@{}}Number of Users for testing \\ (unseen data)\end{tabular}} & \textbf{\begin{tabular}[c]{@{}c@{}}Number of Training\\ Signature Samples\end{tabular}} & \textbf{\begin{tabular}[c]{@{}c@{}}Number of Testing\\ Signature Samples\end{tabular}} & \textbf{Accuracy(\%)} \\ \hline \hline
95                                                                                           & 05                                                                                            & 28500                                                                                   & 1500                                                                                   & 100.00                 \\ 
90                                                                                           & 10                                                                                            & 27000                                                                                   & 3000                                                                                   & 100.00                \\ 
80                                                                                           & 20                                                                                            & 24000                                                                                   & 6000                                                                                  & 100.00                 \\ 
70                                                                                           & 30                                                                                            & 21000                                                                                   & 9000                                                                                  & 100.00                 \\ 
60                                                                                           & 40                                                                                            & 18000                                                                                   & 12000                                                                                  & 100.00                 \\ 
50                                                                                           & 50                                                                                            & 15000                                                                                   & 15000                                                                                  & 100.00                  \\ 
40                                                                                           & 60                                                                                            & 12000                                                                                   & 18000                                                                                  & 100.00               \\ 
30                                                                                           & 70                                                                                            & 9000                                                                                   & 21000                                                                                  &  100.00                \\ 
20                                                                                           & 80                                                                                            & 6000                                                                                   & 24000                                                                                  & 100.00               \\ 
10                                                                                           & 90                                                                                            & 3000                                                                                    & 27000                                                                                  &100.00                 \\ 
05                                                                                           & 95                                                                                            & 1500                                                                                    & 28500                                                                                  & 99.96                \\ 
01 (One Shot Learning)                                                                       & 99                                                                                            & 300                                                                                     & 29700                                                                                  &99.62                \\ \hline
\end{tabular}
\end{center}
\label{tab:tab4}
\end{table*}

\begin{table*}[]
\scriptsize
\caption{CLASSIFICATION ACCURACY OF THE PROPOSED FRAMEWORK WITH MCYT-330 DATASET INCLUDING FORGERY SAMPLES.}
\begin{center}
\begin{tabular}{ccccc}
\hline
\textbf{\begin{tabular}[c]{@{}c@{}}Number of Users for training \\ (seen data)\end{tabular}} & \textbf{\begin{tabular}[c]{@{}c@{}}Number of Users for testing\\ (unseen data)\end{tabular}} & \textbf{\begin{tabular}[c]{@{}c@{}}Number of Training \\ Signature Samples\end{tabular}} & \textbf{\begin{tabular}[c]{@{}c@{}}Number of Testing\\ Signature Samples\end{tabular}} & \textbf{Accuracy(\%)} \\ \hline \hline
329                                                                                          & 01                                                                                           & 197400                                                                                   & 600                                                                                    & 96.50                 \\ 
300                                                                                          & 30                                                                                           & 180000                                                                                   & 18000                                                                                  & 93.51                 \\ 
250                                                                                          & 80                                                                                           & 150000                                                                                   & 48000                                                                                  & 90.75                 \\ 
200                                                                                          & 130                                                                                          & 120000                                                                                   & 78000                                                                                  & 89.13                 \\ 
150                                                                                          & 180                                                                                          & 90000                                                                                    & 108000                                                                                 & 87.63                 \\ 
100                                                                                          & 230                                                                                          & 60000                                                                                    & 138000                                                                                 & 87.50                 \\ 
70                                                                                           & 260                                                                                          & 42000                                                                                    & 156000                                                                                 & 87.45                 \\ 
50                                                                                           & 280                                                                                          & 30000                                                                                    & 168000                                                                                 & 86.36                 \\ 
01 ( One shot learning)                                                                      & 329                                                                                          & 600                                                                                      & 197400                                                                                 & 78.89                 \\ \hline
\end{tabular}
\end{center}
\label{tab:tab5}
\end{table*}

\begin{table*}[]
\scriptsize
\caption{CLASSIFICATION ACCURACY OF THE PROPOSED FRAMEWORK WITH MCYT-330 DATASET EXCLUDING FORGERY SAMPLES.}
\begin{center}
\begin{tabular}{ccccc}
\hline
\textbf{\begin{tabular}[c]{@{}c@{}}Number of Users for training \\ (seen data)\end{tabular}} & \textbf{\begin{tabular}[c]{@{}c@{}}Number of Users for testing\\ (unseen data)\end{tabular}} & \textbf{\begin{tabular}[c]{@{}c@{}}Number of Training \\ Signature Samples\end{tabular}} & \textbf{\begin{tabular}[c]{@{}c@{}}Number of Testing\\ Signature Samples\end{tabular}} & \textbf{Accuracy(\%)} \\ \hline \hline
329                                                                                          & 01                                                                                           & 98700                                                                                   & 300                                                                                    & 100.00                 \\ 
300                                                                                          & 30                                                                                           & 90000                                                                                   & 9000                                                                                  & 100.00                 \\ 
250                                                                                          & 80                                                                                           & 75000                                                                                   & 24000                                                                                  & 100.00                 \\ 
200                                                                                          & 130                                                                                          & 60000                                                                                   & 39000                                                                                  & 100.00                 \\ 
150                                                                                          & 180                                                                                          & 45000                                                                                    & 54000                                                                                 & 100.00                 \\ 
100                                                                                          & 230                                                                                          & 30000                                                                                    & 69000                                                                                 & 100.00                 \\ 
70                                                                                           & 260                                                                                          & 21000                                                                                    & 78000                                                                                 & 100.00                 \\ 
50                                                                                           & 280                                                                                          & 15000                                                                                    & 84000                                                                                 &99.78               \\ 
01 ( One shot learning)                                                                      & 329                                                                                          & 300                                                                                      & 98700                                                                                 &87.59                 \\ \hline
\end{tabular}
\end{center}
\label{tab:tab6}
\end{table*}

There are only few frameworks \cite{tolosana2017biometric, ahrabian2017usage, pei2016modeling} have been proposed for OSV based on Siamese networks. Due to inconsistencies in many aspects, a direct comparison between these works is typically not possible because of the differences in the datasets used for evaluation (commercial or free), subsets of the dataset retrieved for training and testing scenarios, number of training and testing signature samples, using forgeries in testing or not, at which level to compare (feature, preprocessing, score or decision, classifier) etc. \cite{yilmaz2018hybrid}. For comparative study, we have considered similar models which are validated based on MCYT data corpus (DB1 and DB2). The reason is that MCYT-100 evaluates the model, in case, where the lesser number of training and testing samples are available. MCYT-330 evaluates the model with larger number of training and testing samples. Out of the frameworks \cite{tolosana2017biometric, ahrabian2017usage, pei2016modeling}, Pei \textit{et al.} \cite{pei2016modeling} considered MCYT-100 for their evaluation, hence, we evaluated and compared the proposed CNN-Siamese based OSV model with the model proposed by Pei \textit{et al.} \cite{pei2016modeling}.
Tolosana \textit{et al.} \cite{tolosana2017biometric} proposed an LSTM based Siamese network using BiosecurID \cite{fierrez2010biosecurid} dataset. BiosecurID consists of signatures of 400 users, which consists of 16 original signatures and 12 skilled forgeries per user. Therefore, a total of 120 genuine-genuine and 192 genuine-forgery combinations per user. Therefore, a total of 48000 genuine-genuine and genuine-forgery combinations are available in entire dataset. 
Ahrabian \textit{et al.} \cite{ahrabian2017usage} evaluated their Siamese based OSV model using SigWiComp2013-Japanese dataset \cite{malik2013icdar} which contains signatures of 31 users, GPDSsyntheticOnLineOnLineSignature dataset \cite{ferrer2017behavioral} with 1000 users. 

Table \ref{tab:tab3} and \ref{tab:tab4} demonstrates the classification accuracy of the proposed OSV framework on MCYT-100 dataset. Table \ref{tab:tab3} summarizes how the classification accuracy varies in case of including both the genuine - forgery signature pairs for testing the framework. Table \ref{tab:tab4} summarizes the accuracy of framework in which only genuine-genuine combination is used for testing and genuine-forgery combination is not considered. As the number of users considered for training increases, the classification accuracy increases. In case of one-shot learning i.e. considering only one user signature samples for training and testing with remaining users signature samples, achieved the best results of 78.16\% and 99.62\% of classification accuracies. This is quite realistic as forgery samples are removed, automatically the classification accuracy increases.

Table \ref{tab:tab5} and \ref{tab:tab6} demonstrates the classification accuracy of the proposed Siamese network based OSV framework on MCYT-330 dataset. MCYT-330 dataset consists of a total of 198000 genuine-genuine and genuine-forgery combinations. Evaluating the model with vast number of samples and achieving the better classification accuracy confirms the efficiency of the proposed Siamese network based OSV framework.

Table \ref{tab:tab7}-\ref{tab:tab8} demonstrates the classification accuracies in case of SVC dataset w.r.t to number of user’s signature considered for training. The classification accuracies achieved by the proposed framework in case of SVC dataset is less compared to MCYT-100 and MCYT-330 datasets. This is perfectly valid and justifiable, due to the fact that the framework trained on a comparatively larger and diverse dataset is more robust and learns the representational features effectively. Also, the signature datasets deliver varied performances with same protocol, as they differ in acquisition process, devices used for acquisition etc. As illustrated in Table \ref{tab:tab2}, in case of SVC dataset, the number of users, number of features, number genuine and forgery samples, number of features, genuine-genuine and genuine-forgery signature combinations are very less compared to MCYT-100 and MCYT-330. Due to lesser number of training data, the framework may not learn the parameters efficiently. 

\begin{table*}[]
\scriptsize
\caption{CLASSIFICATION ACCURACY OF THE PROPOSED FRAMEWORK WITH SVC DATASET INCLUDING FORGERY SAMPLES.}
\begin{center}
\begin{tabular}{ccccc}
\hline
\textbf{\begin{tabular}[c]{@{}c@{}}Number of Users for training \\ (seen data)\end{tabular}} & \textbf{\begin{tabular}[c]{@{}c@{}}Number of Users for testing \\ (unseen data)\end{tabular}} & \textbf{\begin{tabular}[c]{@{}c@{}}Number of Training \\ Signature Samples\end{tabular}} & \textbf{\begin{tabular}[c]{@{}c@{}}Number of Testing\\ Signature Samples\end{tabular}} & \textbf{Accuracy(\%)} \\ \hline \hline
35                                                                                           & 05                                                                                            & 13300                                                                                    & 1900                                                                                   & 77.00                 \\ 
30                                                                                           & 10                                                                                            & 11400                                                                                    & 3800                                                                                   & 68.21                 \\ 
20                                                                                           & 20                                                                                            & 7600                                                                                     & 7600                                                                                   & 63.95                 \\ 
10                                                                                           & 30                                                                                            & 3800                                                                                     & 11400                                                                                  & 66.65                 \\ 
05                                                                                           & 35                                                                                            & 1900                                                                                     & 13300                                                                                  & 68.63                 \\ 
01 (One shot learning)                                                                       & 39                                                                                            & 380                                                                                      & 14820                                                                                  & 50.00                 \\ \hline
\end{tabular}
\end{center}
\label{tab:tab7}
\end{table*}

\begin{table*}[]
\scriptsize
\caption{CLASSIFICATION ACCURACY OF THE PROPOSED FRAMEWORK WITH SVC DATASET EXCLUDING FORGERY SAMPLES.}
\begin{center}
\begin{tabular}{ccccc}
\hline
\textbf{\begin{tabular}[c]{@{}c@{}}Number of Users for training \\ (seen data)\end{tabular}} & \textbf{\begin{tabular}[c]{@{}c@{}}Number of Users for testing \\ (unseen data)\end{tabular}} & \textbf{\begin{tabular}[c]{@{}c@{}}Number of Training \\ Signature Samples\end{tabular}} & \textbf{\begin{tabular}[c]{@{}c@{}}Number of Testing\\ Signature Samples\end{tabular}} & \textbf{Accuracy(\%)} \\ \hline \hline
35                                                                                           & 05                                                                                            & 6650                                                                                    & 950                                                                                   & 100.00                 \\ 
30                                                                                           & 10                                                                                            & 5700                                                                                    & 1900                                                                                   & 100.00                 \\ 
20                                                                                           & 20                                                                                            & 3800                                                                                     & 3800                                                                                   & 99.50                \\ 
10                                                                                           & 30                                                                                            & 1900                                                                                     & 5700                                                                                  & 99.02                 \\ 
05                                                                                           & 35                                                                                            & 950                                                                                     & 6650                                                                                  & 78.45                 \\ 
01 (One shot learning)                                                                       & 39                                                                                            & 190                                                                                      & 7410                                                                                  & 66.98                 \\ \hline
\end{tabular}
\end{center}
\label{tab:tab8}
\end{table*}

Aligned with the core purpose of Siamese network i.e. the ability to learn the representational features from one signature samples of one user, i.e. one-shot learning, the proposed framework achieves the best accuracy compared to the recent models. This proves that the framework has the ability to learn the representational features even from a single user and able to accurately classify the test signature combinations. 
In Table \ref{tab:tab9}, we compare our framework results in case of MCYT-100 and MCYT-330 datasets with the results achieved by the framework proposed by Pei \textit{et al.} \cite{pei2016modeling}, which is the only model which used MCYT-100 for their model evaluation. They evaluated their model by training the genuine-genuine and genuine-forgery combinations of first 70 (1-70) users and tested with the remaining 30 users (71-100). As depicted in tables \ref{tab:tab3}-\ref{tab:tab6}, we have evaluated our model with all the possible training and testing combinations starting from user 1 and gradually moving to user 99 in case of MCYT-100 and 329 in case of MCYT-330. Table \ref{tab:tab7} confirms that we outperformed Pei \textit{et al.} \cite{pei2016modeling} model. 

In literature, even though lots of online signature verification models has been proposed based on SVC dataset \cite{guru2017interval, guru2007symbolic, manjunatha2016online}, no prior work has been reported on using SVC dataset in Siamese network based online signature verification. As discussed above, Siamese networks verifies whether two input signature pairs belong to same category or not, whereas the non Siamese based classification models takes a single signature as input and classifies whether signature is genuine or forgery. Hence, in order to provide fair evaluation we do not provide the comparison analysis of our proposed framework with other traditional classification works w.r.t SVC dataset.

\begin{figure*}[h]
       \centering
       \fbox{
       \includegraphics[scale=0.6]{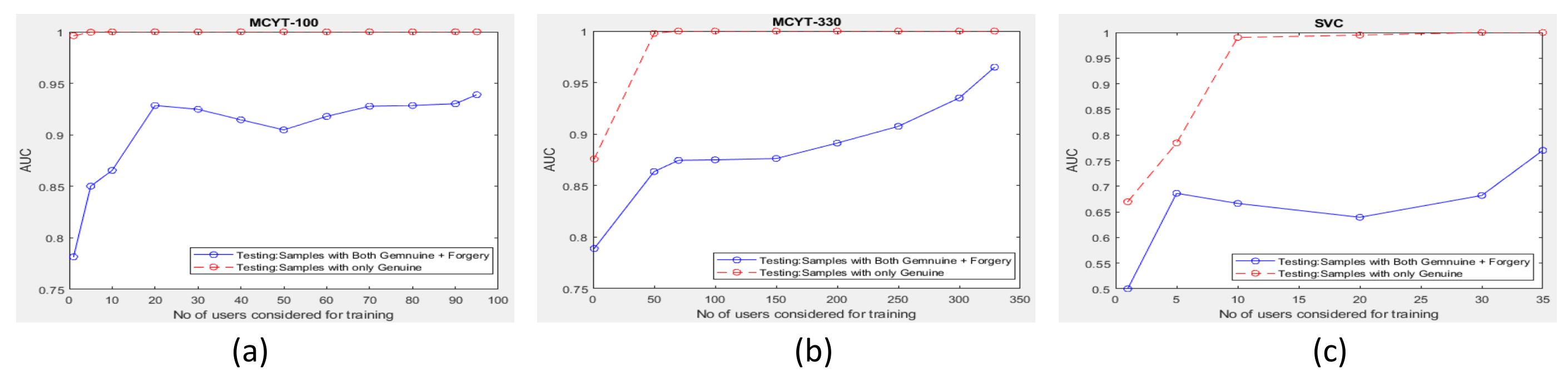}}
       \caption{Area under the receiving-operator curve (AUC) of two test scenarios of proposed framework with (a) MCYT-100, (b) MCYT-330 and (c) SVC datasets}
       \label{fig:graph}
\end{figure*}

\begin{table}[]
\scriptsize		
\caption{COMPARATIVE ANALYSIS OF THE PROPOSED MODEL AGAINST THE RECENT MODELS ON MCYT-100 AND MCYT-330 DATASET. \cite{pei2016modeling}: considered only one case for experimentation: 70 users signature for training and remaining 30 users signature for testing. '-' indicates that the values have not been computed in the respective papers.}
\begin{center}
\begin{tabular}{ccccc}
\hline
\multirow{2}{*}{\textbf{Method}}                                                                                           & \multicolumn{2}{c}{\textbf{MCYT-100}} & \multicolumn{2}{c}{\textbf{MCYT-330}} \\ \cline{2-5} 
                                                                                                                           & \textbf{01}       & \textbf{70}       & \textbf{01}       & \textbf{70}       \\ \hline \hline
\textbf{\begin{tabular}[c]{@{}c@{}}Proposed Model – \\ (CNN+ Siamese)\end{tabular}}                                        & 78.6              & 92.7              & 100.0             & 100.0             \\ 
\textbf{\begin{tabular}[c]{@{}c@{}}Siamese+Recurrent Network – \\ Average (without forgery) {[}26{]}\end{tabular}}         & -                 & 91.4              & -                 & -                 \\
\textbf{\begin{tabular}[c]{@{}c@{}}Siamese+Recurrent Network -\\  Last timestep (without forgery) {[}26{]}\end{tabular}}   & -                 & 92.0              & -                 & -                 \\ 
\textbf{\begin{tabular}[c]{@{}c@{}}Siamese Network-\\ Average (without forgery) {[}26{]}\end{tabular}}                     & -                 & 81.6              & -                 & -                 \\ 
\textbf{\begin{tabular}[c]{@{}c@{}}Siamese Network-\\  Last timestep (without forgery) {[}26{]}\end{tabular}}              & -                 & 76.0              & -                 & -                 \\ 
\textbf{\begin{tabular}[c]{@{}c@{}}Siamese+Recurrent Network –\\  Average (including forgery) {[}26{]}\end{tabular}}       & -                 & 88.8              & -                 & -                 \\ \textbf{\begin{tabular}[c]{@{}c@{}}Siamese+Recurrent Network - \\ Last timestep (including forgery) {[}26{]}\end{tabular}} & -                 & 87.6              & -                 & -                 \\ 
\textbf{\begin{tabular}[c]{@{}c@{}}Siamese Network-\\ Average (including forgery) {[}26{]}\end{tabular}}                   & -                 & 82.8              & -                 & -                 \\
\textbf{\begin{tabular}[c]{@{}c@{}}Siamese Network- \\ Last timestep (including forgery) {[}26{]}\end{tabular}}            & -                 & 66.8              & -                 & -                 \\ \hline
\end{tabular}
\end{center}
\label{tab:tab9}
\end{table}

Fig. \ref{fig:graph}, illustrates the Receiving-Operator Curves (ROC) of the proposed model, in both the test scenarios, i.e. both `genuine plus forgery' and `only genuine'. In case of MCYT-100, Area under the receiving-operator curve (AUC) in case of `genuine and forgery' combination achieves a value close to 0.95 and the trend increases with the increase of number of users considered for training. In case of `only genuine', AUC reaches the peaks by training even with only one user. Similar trend is exhibited by the MCYT-330 and SVC datasets. 
To conclude this section, we see that our proposed Siamese network based OSV framework, effectively learns to place the similar pairs in proximity and dissimilar pairs far in embedded feature space. The model excels in true application of Siamese network i.e. one-shot based learning by achieving start-of-the results in all the datasets. Although the proposed framework achieved efficient results, the tables summarizes that there is a scope for improvement in case of fewer number of user’s signature samples used for training samples.

\section{CONCLUSION AND FUTURE WORK}
\label{sec:conclusion}
In this paper, we presented a novel CNN - Siamese based writer-independent OSV frame work to address the two most challenging co-variates of online signature verification i.e. one-shot learning and accurately learn the intra-personal variations of the signature. Extensive experiments are demonstrated to evaluate the CNN-Siamese based models on both large and small datasets i.e. MCYT-330 and SVC. The high accuracy in case of SVC dataset confirms that the better representational ability of the proposed model to deliver high classification accuracies even with less data and best suited for real time  applications. In contrast to the recent Siamese network based models, we have thoroughly evaluated the proposed model by testing the model with varying number of samples. The  proposed model demonstrated to be capable of achieving excellent performance by surpassing the recent state-of-the-art baseline models.




\bibliographystyle{IEEEtran}
\bibliography{reference}

\end{document}